\documentclass[final]{cvpr}

\usepackage{times}
\usepackage{epsfig}
\usepackage{graphicx}
\usepackage{amsmath}
\usepackage{amssymb}
\usepackage{xcolor}

\usepackage{booktabs}
\usepackage[font={small}]{caption}

\usepackage[shrink=10]{microtype} %

\usepackage[pagebackref=true,breaklinks=true,colorlinks,bookmarks=false]{hyperref}
\usepackage[capitalise]{cleveref}

\usepackage{abstract}
\def\abstract
   {%
   \centerline{\large\bf Abstract}%
   \vspace*{12pt}%
   \it%
   }

\newcommand{\myparagraph}[1]{\smallskip\noindent\textbf{#1}}
\newcommand{\SAD}{SAD}
\newcommand{\ourmodel}{\textbf{\textit{RADAR}}\@\xspace}

\def\cvprPaperID{340} %
\def\confYear{CVPR 2021}
\title{De-rendering the World's Revolutionary Artefacts}
\author{Shangzhe Wu$^{1,4}$\thanks{The work was primarily done during an internship at Google Research.}
\qquad
Ameesh Makadia$^4$
\qquad
Jiajun Wu$^2$ \\
Noah Snavely$^4$
\qquad
Richard Tucker$^4$
\qquad
Angjoo Kanazawa$^{3,4}$
\vspace{0.5em} \\
$^1$University of Oxford \qquad
$^2$Stanford University \qquad \\
$^3$University of California, Berkeley \qquad
$^4$Google Research
}

\begin{document}

\twocolumn[
\maketitle
\vspace{-3em}
\begin{center}
  \includegraphics[width=1\linewidth]{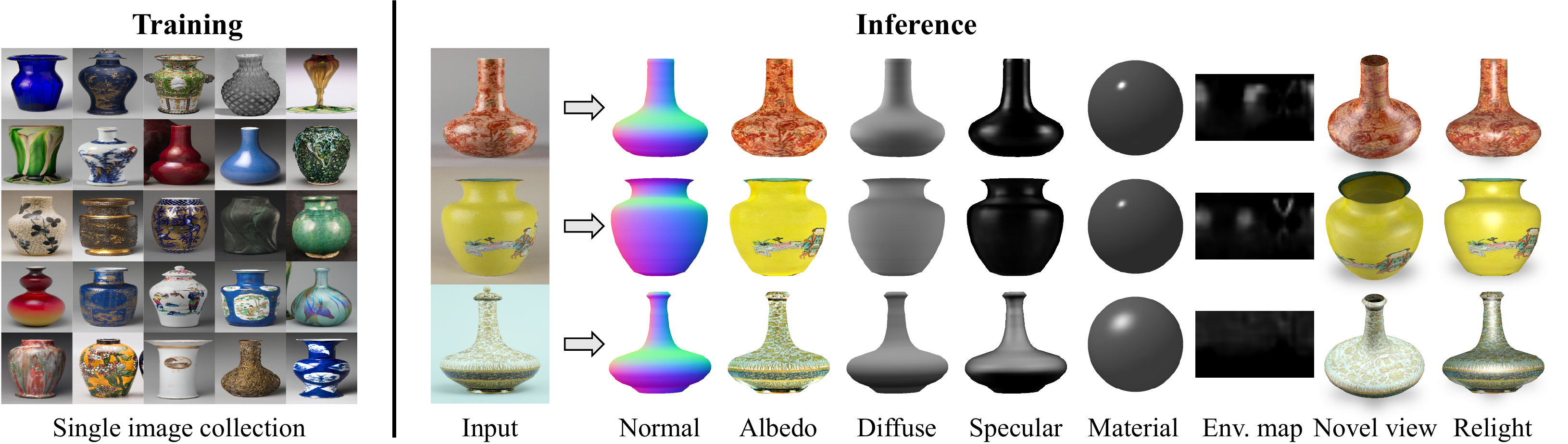}
\captionof{figure}{\textbf{De-rendering from single images.} %
From only a real single-view image collection of ``revolutionary'' (i.e., solid of revolution) artefacts with known silhouettes as training data (left), our framework learns to de-render a single image into shape, albedo and complex lighting and material components, suitable for applications such as novel-view synthesis and relighting (right). 
}
\label{fig:teaser}
\end{center}

\medbreak
]

\saythanks  %
\begin{abstract}
Recent works have shown exciting results in unsupervised image de-rendering---learning to decompose 3D shape, appearance, and lighting from single-image collections without explicit supervision. 
However, many of these assume simplistic material and lighting models.
We propose a method, termed RADAR, that can recover environment illumination and surface materials from real single-image collections,
relying neither on explicit 3D supervision, nor on multi-view or multi-light images. Specifically, we focus on rotationally symmetric artefacts that exhibit challenging surface properties including specular reflections, such as vases.
We introduce a novel self-supervised albedo discriminator, which allows the model to recover plausible albedo without requiring any ground-truth during training.
In conjunction with a shape reconstruction module exploiting rotational symmetry, we present an end-to-end learning framework that is able to de-render the world's revolutionary artefacts.
We conduct experiments on a real vase dataset and demonstrate compelling decomposition results, allowing for applications including free-viewpoint rendering and relighting.
More results and code at: {\footnotesize\url{https://sorderender.github.io/}}.
\end{abstract}

\section{Introduction}

Consider one of the vases shown in \cref{fig:teaser}. From just a single image, we can tell a lot about the underlying properties of that vase. Despite the image's flatness, we can perceive 
an instance of a 3D surface with various lights cast upon it.
We can distinguish between areas where the underlying color of the vase changes and regions that reflect light, revealing the glossiness of the surface and its local geometry.
We introduce a model that aims to \emph{de-render} a single image into these 
factors---geometry, material, and illumination---which we call \ourmodel (\textbf{\textit{R}}evolutionary \textbf{\textit{A}}rtefact \textbf{\textit{D}}e-rendering \textbf{\textit{A}}nd \textbf{\textit{R}}e-rendering).
In particular, our approach can decompose real images of vase-like objects under complex illumination and with glossy materials.
Notably, our approach can learn this ability just from collections of single images (i.e., where each object is pictured once), without explicit 3D supervision or multiple images.
This allows us to analyze images obtained in real world settings, such as artefact collections in museums, and subsequently apply modifications including relighting, as illustrated in \cref{fig:teaser}.

Making de-rendering tractable
involves
simplifying assumptions. In some methods, this means requiring explicit supervision,
e.g., with synthetic~\cite{li2020inverse,li2018learning} or specially captured data~\cite{li2018materials}. An alternative to direct supervision is to observe an object under multiple viewpoints~\cite{bi2020deep,yariv2020multiview} or multiple lights~\cite{xu2019deep,Boss2020brdf},
but for many existing image collections,
such multiple views are unavailable. 
Hence, 
learning to de-render from single image collections has been of growing interest~\cite{Wu_2020_CVPR,sahasrabudhe2019lifting}. %
However, these approaches assume simplistic shading or lighting models, such as Lambertian,
and are not applicable to realistic 
scenarios with complex illumination effects. 

In contrast, our objective in this paper is to explore unsupervised de-rendering in the presence of more complex illumination effects.
To make our task tractable, we  consider simplifying assumptions on the 3D shape.
We draw inspiration from recent work~\cite{Wu_2020_CVPR} that leverages symmetry priors for self-supervised decomposition.
Specifically, we focus on de-rendering objects whose shapes are described by solids of revolution (SoRs, or ``\emph{revolutionary}'' objects)---such objects include many categories of man-made objects such as vases.
This allows us to derive a simple yet effective method for recovering the 3D geometry and camera viewpoint from only single images with 2D silhouettes.

Our model de-renders a single image of a revolutionary object into 3D geometry, viewpoint, albedo, material shininess, and environment lighting. Even with this strong assumption on SoR shape and inductive bias on the rendering process, this is still an extremely under-constrained problem. As with most ill-posed inverse problems, we must prevent degenerate solutions where the model learns no disentanglement at all. Another major challenge is to predict realistic diffuse albedo in regions saturated by specular reflections.
To ensure realistic disentanglement, we incorporate novel components into our model. In particular, we propose a new adversarial module that we call a \emph{Self-supervised Albedo Discriminator} (\SAD). The key insight is that the distribution of diffuse albedo patches should be independent of observed specular effects---it should not be possible to tell from the albedo alone whether a particular surface region exhibits a specular reflection or not. 
Unlike existing adversarial frameworks, a key feature %
of SAD is that the discriminator always takes its inputs from the predicted albedo and never requires a `real' albedo, hence the label \emph{self-supervised}. 
In summary, we propose \ourmodel, an end-to-end framework for de-rendering single images into shape, complex lighting, and
materials, learning only from single-image collections with 2D silhouettes.
We 
evaluate our approach numerically on a synthetic dataset, and demonstrate effective results on real images of revolutionary artefacts from museum collections, where our approach allows for applications such as free-viewpoint rendering and relighting.

\begin{figure*}
\includegraphics[width=1\linewidth]{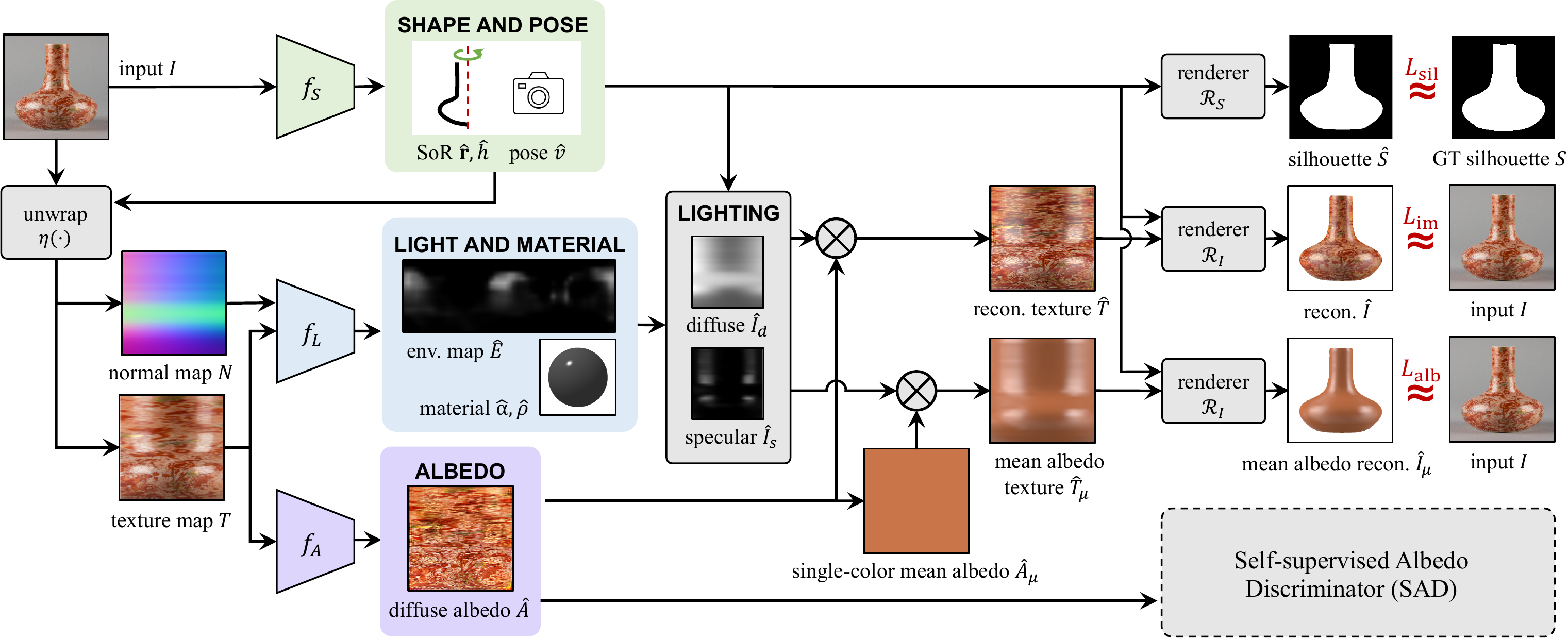}
\caption{\textbf{\ourmodel training pipeline.} Given a single image of a vase, our model first predicts shape and pose with the shape network $f_S$, which is used to unwrap the surface of the object.
The lighting network $f_L$ and albedo network $f_A$ then take in the unwrapped textures and predict environment lighting, surface material and diffuse albedo, which are recomposed to render the input image.
A self-supervised albedo discriminator is proposed to encourage the decomposition of albedo and lighting, illustrated in~\cref{fig:albedo_disc}.
The whole pipeline is trained end-to-end without any external supervision except for the silhouettes.}
\label{fig:pipeline}
\vspace{-0.5em}
\end{figure*}

\section{Related Work}
There is a vast literature on intrinsic image decomposition and de-rendering. Many methods build upon some physical model of the image formation process and complement such models with representations learned from data.
Existing methods can be roughly divided into three categories: optimization-based, learning from annotated or synthetic images, and learning from unannotated image collections. We focus 
on single-image decomposition methods.

\myparagraph{Optimization-based approaches.} Traditional approaches derive heuristic physical priors and rely on optimization with such priors to decompose images~\cite{horn1975obtaining,barrow1978computer,land1971lightness,BarronTPAMI2015,goldman2009shape}. In particular, SIRFS~\cite{BarronTPAMI2015} is an extension of classic shape-from-shading that recovers shape as well as reflectance and illumination, but does not handle non-Lambertian reflectance.
While these methods work well in specific domains, it turns out to be challenging to design general priors for real images with complex intrinsic albedo and BRDFs.

\myparagraph{Learning from annotated images.}
Leveraging advances in deep learning, researchers have explored learning-based
intrinsic image decomposition.
Shi~\etal~\cite{ShapenetIntrinsics}
use synthetic ShapeNet objects for training; Liu~\etal\cite{liu2017material} extend this framework for material editing.
Others attempt to decompose general objects under flash illumination~\cite{li2018learning, sang2020single} or general indoor scenes~\cite{li2020inverse}, similarly with synthetic data.
However, models trained purely only synthetic data often generalize poorly to real scenes due to the domain gap.
Some methods pre-train models on synthetic data and then fine-tune them on real data~\cite{janner2017intrinsic, sfsnetSengupta18} for better generalization.
Tremendous efforts are still required to generate large-scale realistic synthetic data that allows easy fine-tuning.

A few works have also studied learning from controlled data, such as multi-view or multi-light images. Many of them also require multiple images during inference~\cite{xu2019deep, bi2020deep, yariv2020multiview, Boss2020brdf}.
Kulkarni~\etal~\cite{kulkarni2015deep} and Ma~\etal~\cite{ma2018single} leverage training pipelines that allow for single-image inference.
However, the complexity in acquiring controlled multiple images of the same real-world object has led these models to be trained again only on synthetic data.
Some recent works leverage photo collections of real scenes~\cite{laffont2012coherent,yu19inverserendernet,YuSelfRelight20,liu2020factorize}, but are often restricted to famous landmarks or street view imagery.

\myparagraph{Learning from unannotated image collections.}
As explicit or indirect supervision 
is rarely available for real-world objects and synthetic datasets often lack sufficient realism, a few recent papers have attempted to learn image decomposition directly from unannotated real image collections~\cite{cmrKanazawa18,chen2019dibrender,Wu_2020_CVPR,sahasrabudhe2019lifting}, but none of them can recover complex material and lighting effects, such as specular reflection.

Our method follows a similar setup, and is able to recover environment illumination and glossy material properties from a single image.
Inspired by Wu~\etal~\cite{Wu_2020_CVPR}, which leverages a bilateral symmetry assumption to recover shape, albedo and diffuse lighting, our model also embraces a rotational symmetry prior to obtain the shape of sufficient quality,
allowing us to start to reason about complex material and illumination in real images.
Rotational symmetry has been exploited for shape recovery in prior work~\cite{colombo05pami,chen2013sweep,Phillips2016SeeingGF,Chen_2020}, but our method also recovers material and lighting. %
\section{Method}

\begin{figure}
\includegraphics[width=1\linewidth]{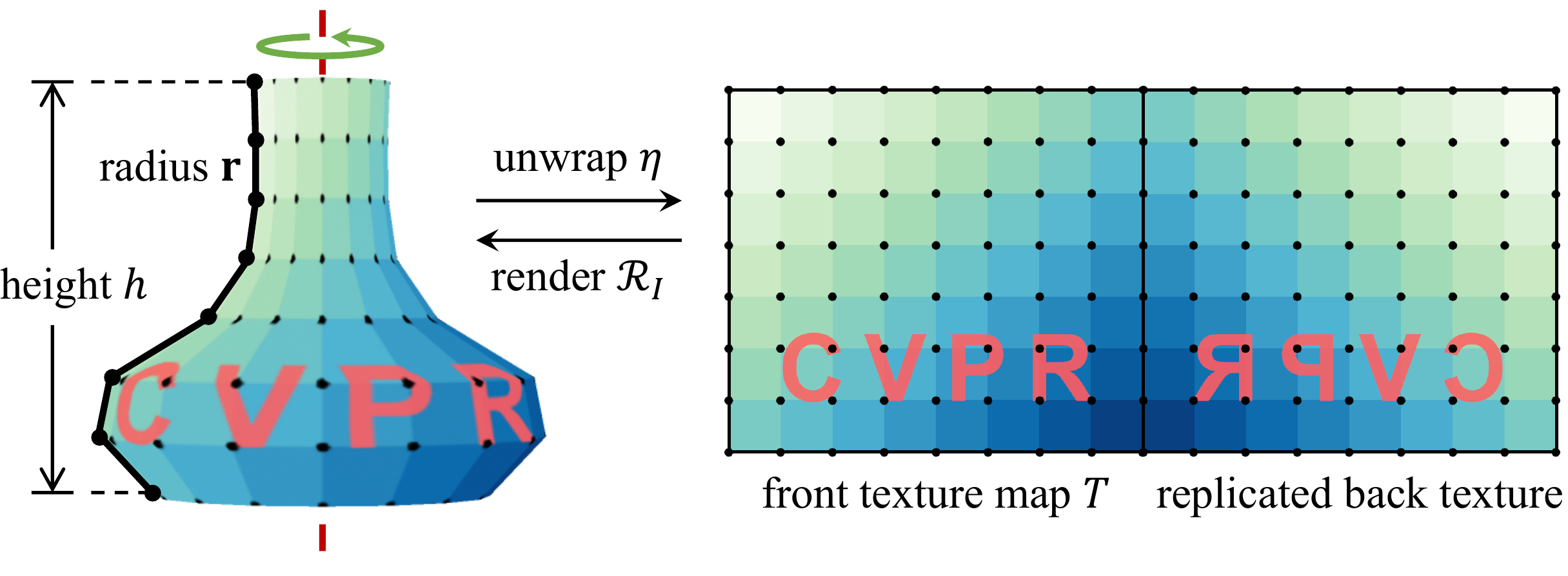}
\vspace{-2em}
\caption{\textbf{Surface of revolution.} We represent the surface using a vertex grid $V \in \mathbb{R}^{L \times K \times 3}$ generated by rotating a discretized radius curve $\mathbf{r} \in \mathbb{R}^L$ around the axis of revolution.}
\label{fig:sor}
\vspace{-0.5em}
\end{figure}

Given a collection of single-view images of revolutionary artefacts, such as vases, our goal is to learn a de-rendering function $\Phi$, which takes in a single image $I$ and predicts the 3D shape of the object, its surface material properties, and the environment illumination.
Making this even more challenging, we do not want to rely on explicit 3D supervision or multi-view images, as obtaining such supervision is not only expensive but often simply intractable for precaptured image collections.

In general, recovering shape, material, and lighting without direct supervision is an extremely ill-posed inverse problem.
In this paper, we consider this task for objects whose shapes roughly observe \emph{solids of revolution (SoRs)}, and assume that only minimal indirect training supervision is available, in the form of reasonable silhouettes which can be obtained using off-the-shelf object detectors. SoRs describe a large subset of real world objects. In particular, we focus on vases, which are made of materials exhibiting complex lighting effects such as specular reflections.

\cref{fig:pipeline} shows an overview of our training pipeline. In the following sections, we present the main components of model, including three sub-networks that recover the shape ($f_S$), lighting ($f_L$), and diffuse albedo ($f_A$) from a single image, along with the reconstruction losses used to train them. We then describe additional components we introduce to encourage realistic disentanglement of lighting and albedo.

\subsection{SoR Shape and Texture}
Vases made on spinning wheels have `revolutionary' shapes known as solids of revolution (SoRs).
SoRs are generated by a plane curve (the generatrix) rotated about a straight line (the axis of revolution).
We model the shape of a vase as an SoR, parameterized by a vector $\mathbf{r} \in \mathbb{R}^L$ giving the radius (i.e.~the perpendicular distance from the axis to the generatrix) at $L$ evenly-spaced points along the axis of revolution, together with the axis height $h$, as illustrated in \cref{fig:sor}.

A complete discretization of the SoR shape is obtained by rotating the resulting sampled curve $\mathbf{r}$ about the axis to obtain sample points at $K$ evenly-spaced rotation angles in $[0^\circ, 360^\circ)$. This produces a regular sampling of the surface in height and angle, and we denote the resulting vertex map as $V \in \mathbb{R}^{L \times K \times 3}$.
To recover the shape from a single image, we define a shape network $f_S$, which takes an image $I$ and predicts the radius column $\mathbf{\hat{r}}$ and its height $\hat{h}$, as well as the camera pose $\hat{v} \in \mathbb{R}^4$, which specifies pitch and roll Euler angles and translation in the $X$ and $Y$ axes:
\begin{equation}
(\mathbf{\hat{r}}, \hat{h}, \hat{v}) = f_S(I).
\end{equation}
As described above, the vertex map $\hat{V}$ can be constructed from the predicted radius column $\mathbf{\hat{r}}$ and height $\hat{h}$. We use a differentiable renderer $\mathcal{R}_S$~\cite{kato2018renderer} to render a silhouette of the predicted SoR mesh with vertices $\hat{V}$ and camera pose $\hat{v}$:
\begin{equation}
\hat{S} = \mathcal{R}_S (\hat{V}, \hat{v}).
\end{equation}
For simplicity, we fix the camera intrinsics for all rendering operations in our model.
We can train the network by minimizing the silhouette loss:
\begin{equation}
L_{\text{sil}} = \lambda_{\text{s}} \|S - \hat{S}\|_2^2 + \lambda_{\text{dt}} \|\mathtt{dt}(S) \odot \hat{S}\|_1,
\end{equation}
where $\odot$ is the Hadamard product and $S$ the target silhouette, obtained with an off-the-shelf object segmentation technique (details in Sec.~\ref{sec:setup}). $\mathtt{dt}(\cdot)$ is the distance transform of the mask, and $\lambda_\text{s}$, $\lambda_\text{dt}$ are weights balancing the terms.

\myparagraph{Texture representation and unwrapping.}
As described above, our SoR representation allows us to unwrap the surface into a regular 2D grid, which can be easily triangulated for rendering.
To render textures, we define a 2D texture map $T \in \mathbb{R}^{H_T \times W_T \times 3}$ in \emph{the unwrapped space} aligned with the vertex map, which is interpolated and mapped onto the surface during rendering with a differentiable renderer $\mathcal{R}_I$:
\begin{equation}
I = \mathcal{R}_I (V, v, T).
\end{equation}
We denote by $\eta$ the inverse mapping of this texture rendering operation, which unwraps the textures of a SoR surface into a texture map $T$ from an image $I$:
\begin{equation}\label{eq:unwrap}
T = \eta(V, v, I).
\end{equation}
As explained in the next section, we decompose material and lighting in this unwrapped space, since a 2D convolution on the unwrapped texture map will behave closer to an intrinsic convolution on the SoR surface, and it is also viewpoint and shape invariant.
See \cref{fig:sor} for an illustration.
\subsection{Unsupervised De-rendering}
We first describe our lighting model, then the network architecture that can produces these components. Because this is still ill-posed, we discuss our methods to encourage proper disentaglement without explicit supervision.

\myparagraph{Lighting model.}
We use a Phong illumination model~\cite{phong1975} with a normalized specular term and a single-channel environment map $E \in \mathbb{R}^{H_E \times W_E}$ to represent the environment lighting.
Each vase is modeled as having a diffuse albedo texture $A \in \mathbb{R}^{H_T \times W_T \times 3}$ in the unwrapped space, a constant shininess scalar $\alpha$, and a constant specular albedo scalar $\rho$, since specular reflections are often due to a layer of glaze on the vase's surface.
Note that for simplicity, we assume gray illumination leaving the color information to the albedo, and ignore global illumination.
The rendered texture $T$ is given by the tone-mapped sum of diffuse and specular terms:
\begin{equation}\label{eq:texturefull}
T = \tau(A \odot I_d + \rho I_s),
\end{equation}
where the tone-mapper is the inverse gamma function $\tau(I) = I^{1/\gamma}$ with $\gamma = 2.2$.
$I_d, I_s \in\mathbb{R}^{H_T\times W_T}$ are the diffuse and specular lighting factors
also in the unwrapped space, which are computed as follows.

We treat each pixel $i$ in $E$ as a directional light with intensity $I_i$ and lighting direction $\vec{L_i}$. Since $E$ is an equirectangular projection of the sphere, $\vec{L_i}$ can be determined directly from the pixel coordinates.  $I_d$ and $I_s$ are then given by\pagebreak[1]
\begin{align}\label{eq:diffuse}
{I_d}_j &= \sum_{i \in E} I_i (\vec{L_i} \cdot \vec{N}_j), \\
{I_s}_j &= \frac{\alpha+1}{2\pi} \sum_{i \in E} I_i (\vec{R}_{i,j} \cdot \vec{P_j})^{\alpha}. \label{eq:specular}
\end{align}
Here the subscript $j$ denotes a pixel in the unwrapped space, $N$ is the surface normal map, and $P$ the view direction map. Finally, $\vec{R}_{i,j}$ is the reflected light direction computed from the environment light direction at pixel $i$ and the surface normal at pixel $j$: $\vec{R}_{i,j} = 2(\vec{L_i} \cdot \vec{N_j})\vec{N_j} - \vec{L_i}$.
Inspired by energy-conserving Phong models~\cite{Arvo1995}, we include a normalization term in the specular component, which essentially ensures that the cosine lobe integrates to a constant.
We find it helpful during training in preventing the specular
component from vanishing when the shininess $\alpha$ gets large.

\begin{figure}
\includegraphics[width=1\linewidth]{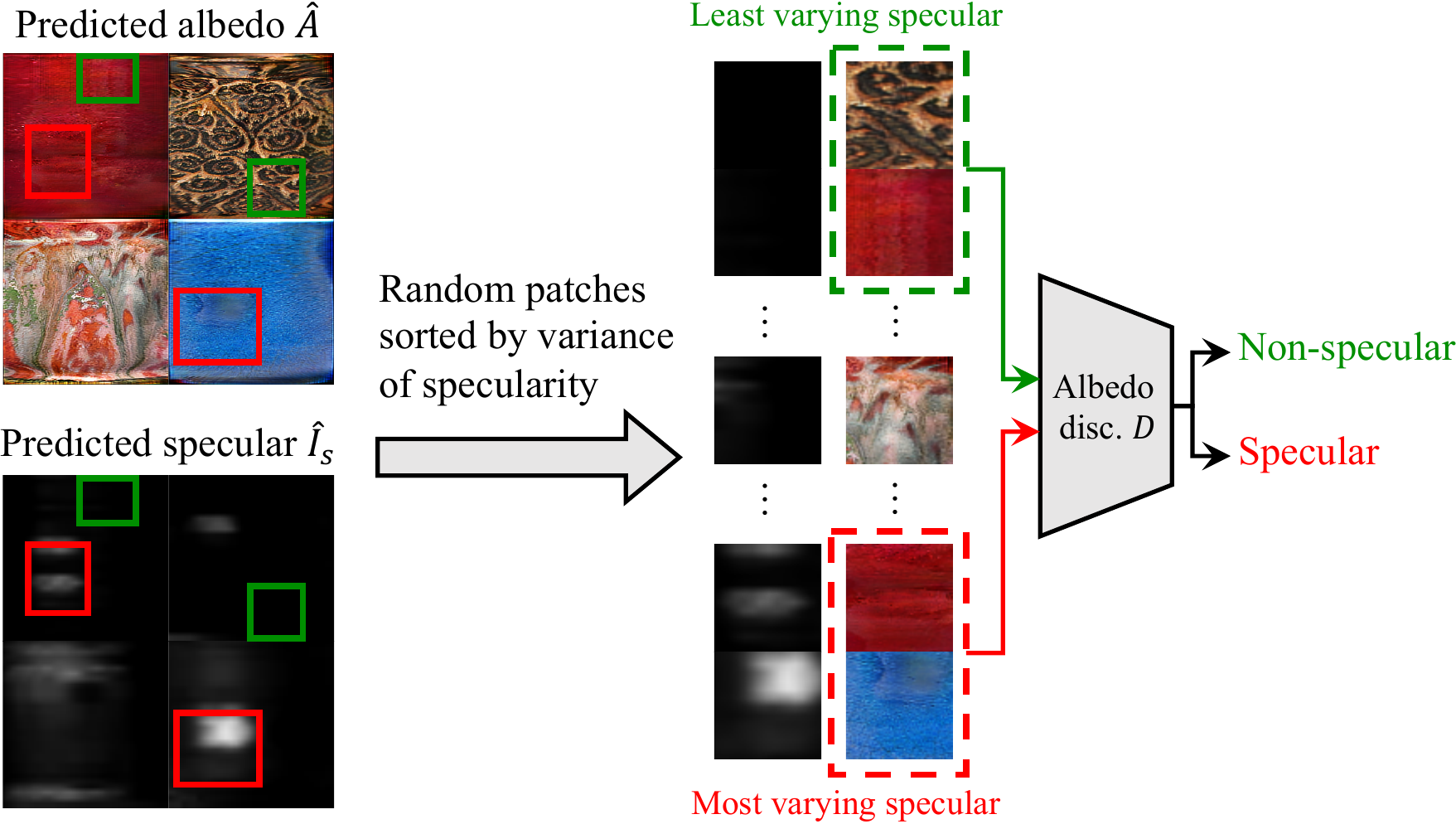}
\caption{\textbf{Self-supervised albedo discriminator.}
We randomly sample patches from the predicted albedo map and sort them by the variance of the corresponding specular patches.
We feed the two groups of albedo patches with the lowest and the highest specular variance to a discriminator, and train our model to prevent it from telling the two groups apart.}
\label{fig:albedo_disc}
\vspace{-0.5em}
\end{figure}

\myparagraph{Predicting materials and lighting.}
Recall that our objective is to de-render an image into various components, specified by $V, v, A, E, \alpha$, and $\rho$. Our model takes the predicted shape and camera pose ($\hat{V}$ and $\hat{v}$) as inputs when recovering the remaining terms (see \cref{fig:pipeline}).

In order to decompose surface materials and illumination, we first unwrap the texture of the frontal (visible) half of the vase from the input image $I$ into a texture map $T \in \mathbb{R}^{H_T \times W_T \times 3}$ using \cref{eq:unwrap} with the predicted shape and pose: $T = \eta(\hat{V}_{\text{f}}, \hat{v}, I)$, where $\hat{V}_{\text{f}}$ denotes the vertices corresponding to the frontal half of the vase.

We design two networks, $f_A$ and $f_L$, to predict albedo and lighting.
The albedo network takes in the unwrapped texture map $T$ and predicts a diffuse albedo map: $\hat{A} = f_A(T)$.
The lighting network takes in an additional normal map $\hat{N}$ concatenated along channel dimension and predicts the environment map, shininess, and specular albedo: $(\hat{E}, \hat{\alpha}, \hat{\rho}) = f_L(T, \hat{N})$. Note that the surface normals $\hat{N} \in \mathbb{R}^{H_T \times W_T \times 3}$ are computed from the predicted vertices $\hat{V}$ and upsampled.

We then apply~\cref{eq:diffuse,eq:specular} to generate predicted lighting factors $\hat{I_d}, \hat{I_s}$, and render a reconstruction of the input image using the differentiable renderer $\mathcal{R}_I$:
\begin{align}\label{eq:tonemap}
\hat{T} &= \tau(\hat{A} \odot \hat{I}_d + \hat{\rho} \hat{I}_s),\\
\label{eq:render}
\hat{I} &= \mathcal{R}_I (\hat{V}_{\text{f}}, \hat{v}, \hat{T}).
\end{align}
We can then train the networks with a reconstruction loss:
\begin{equation}\label{eq:loss_im}
L_{\text{im}} = \|\tilde{S} \odot (I - \hat{I})\|_1,
\end{equation}
where $\tilde{S}$ is the intersection of the ground-truth silhouette $S$ and the rendered silhouette of the frontal visible part $\hat{S}_{\text{f}}$.

\subsection{Disentangling Lighting and Albedo}
Thus far, we have introduced three networks ($f_S, f_A$, $f_E$) to de-render an image into its shape, material, and lighting components. While the loss $L_{\text{im}}$ ensures these components combine to faithfully reproduce the input image, recovering the individual terms correctly remains underconstrained.

From the lighting model (\cref{eq:texturefull}) we can identify two prominent failure modes when training only with the reconstruction loss. First, the model can always predict little or no specularity and leave all the specular reflections in the albedo map. Second, a non-empty specular map is still insufficient to ensure accurate albedo, as there is no incentive for the model to disentangle these components correctly, or to reconstruct realistic albedo in regions saturated by specularity.
In the following, we introduce additional components to our model to prevent these failure modes.

\myparagraph{Single-color albedo rendering.}
To encourage the model to utilize the lighting components, we replace the predicted diffuse albedo map $\hat{A}$ with a single average color of it $\hat{A}_\mu$, and obtain a second reconstructed image $\hat{I}_\mu$ with this single-color albedo. 
We then define another reconstruction loss $L_{\text{alb}}$ similar to \cref{eq:loss_im}:
\begin{equation}\label{eq:loss_sin}
L_{\text{alb}} = \|\tilde{S} \odot (I - \hat{I}_\mu)\|_1.
\end{equation}
This auxiliary loss encourages the lighting network to make a coarse lighting prediction, such that the reconstructed image rendered with single-color albedo can still recover some color variation in the input image resulting from the lighting alone.
However, this does not guarantee correct lighting or albedo predictions, as the single-color approximation does not reflect the color diversity in real vases, or provide a useful signal to reconstruct albedo in saturated specular regions.
Hence, we address these limitations next.

\myparagraph{Self-supervised albedo discriminator.}
In order to successfully recover the diffuse material, we must incentivize the model to predict a realistic albedo map free of specular effects. This is particularly challenging for large patches saturated by specular reflections, which require an inpainting-like solution to recover the underlying albedo. %
To this end, we propose a novel specularity-guided Self-supervised Albedo Discriminator (\SAD).

Starting with the weak assumption that our model can predict a moderately reasonable specular map, we make two key insights. First, the distribution of patches in the true diffuse albedo is independent of the specular map, i.e., it should not be possible to predict specular reflection from the albedo alone. Second, the accuracy of the predicted albedo for an image patch is generally inversely related to the amount of specularity in the patch. This follows from the observation that where the specularity is low, the input texture map is much closer to the true albedo compared to image patches saturated by specular reflections.

From these observations, it follows that we can improve the albedo prediction in highly specular regions by making their distribution indistinguishable from that of the albedo patches in low specular regions. We realize this idea with an adversarial framework~\cite{goodfellow2014generative}.
As illustrated in~\cref{fig:albedo_disc}, for each iteration of during training, we randomly sample patches from the predicted tone-mapped diffuse albedo maps $\{\tau(\hat{A}^{(i)})\}_{i=1}^B$ in a batch, and separate them into two groups according to the variance of their specularity values: one group $\mathcal{P}_{\text{nonspec}}$ with low specularity variance (``real''), and the other $\mathcal{P}_{\text{spec}}$ with high specularity variance (``fake'').

We then have a discriminator network $D$ that tries to tell apart these two groups of albedo patches, and introduce an additional GAN loss to our decomposition model:
\begin{equation}\label{eq:loss_gan}
L_{\text{SAD}} = \mathbb{E}_{p_1 \in \mathcal{P}_{\text{nonspec}}} [\log D(p_1)]
+ \mathbb{E}_{p_2 \in \mathcal{P}_{\text{spec}}} [\log(1- D(p_2))].
\end{equation}
We label our discriminator as self-supervised, as no ``real'' albedo data is necessary in training our framework.
We show in~\cref{fig:ablation} that SAD significantly improves the quality of the albedo prediction, especially in saturated specular regions.

\begin{figure*}
\includegraphics[width=1\linewidth]{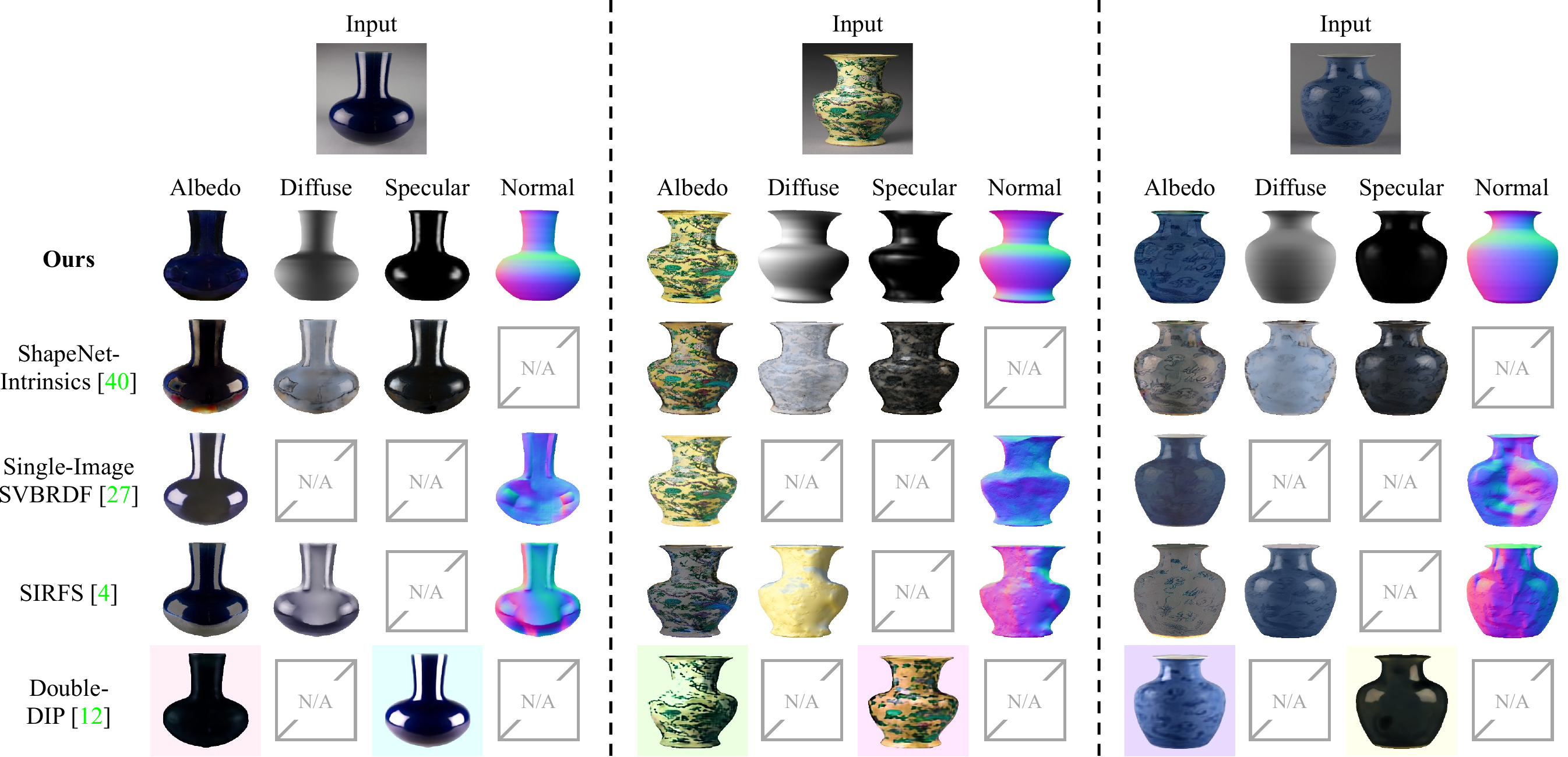}
\caption{\textbf{Qualitative comparison.}
We compare the decomposition results of our model against several prior methods.
Our method recovers accurate geometry and achieves significantly better decomposition compared to other methods, including supervised models trained on synthetic objects or specially captured data~\cite{ShapenetIntrinsics, li2018learning}.
}
\label{fig:compare_real}
\vspace{-0.5em}
\end{figure*}

\subsection{End-to-end Training}
After combining all the components of our model, there remains an inherent ambiguity between the intensity level of the light and the brightness level of the albedo. Thus, we add a consistency regularizer on the diffuse map $\hat{I_d}$, encouraging its average brightness to reside in a specified interval:
\begin{equation}
L_{\text{diff}} = \max\Bigl(\Bigl\lvert\frac{1}{HW} \sum_i \hat{I}_{d,i} - \xi\Bigr\rvert - \Delta, 0\Bigr)^2,
\end{equation}
where $\xi$ specifies the target brightness level and $\Delta$ is the margin, which are respectively set to $0.5$ and $0.1$ in our experiments. The total loss used to train our model end-to-end is a weighted sum of the five loss terms:
\begin{equation}
L_{\text{total}} = L_{\text{sil}} + \lambda_{\text{im}} L_{\text{im}} + \lambda_{\text{alb}} L_{\text{alb}} + \lambda_{\text{SAD}} L_{\text{SAD}} + \lambda_{\text{diff}} L_{\text{diff}}.
\end{equation}

\section{Experiments}

\begin{figure}
\vspace{-1em}
\includegraphics[width=1\linewidth]{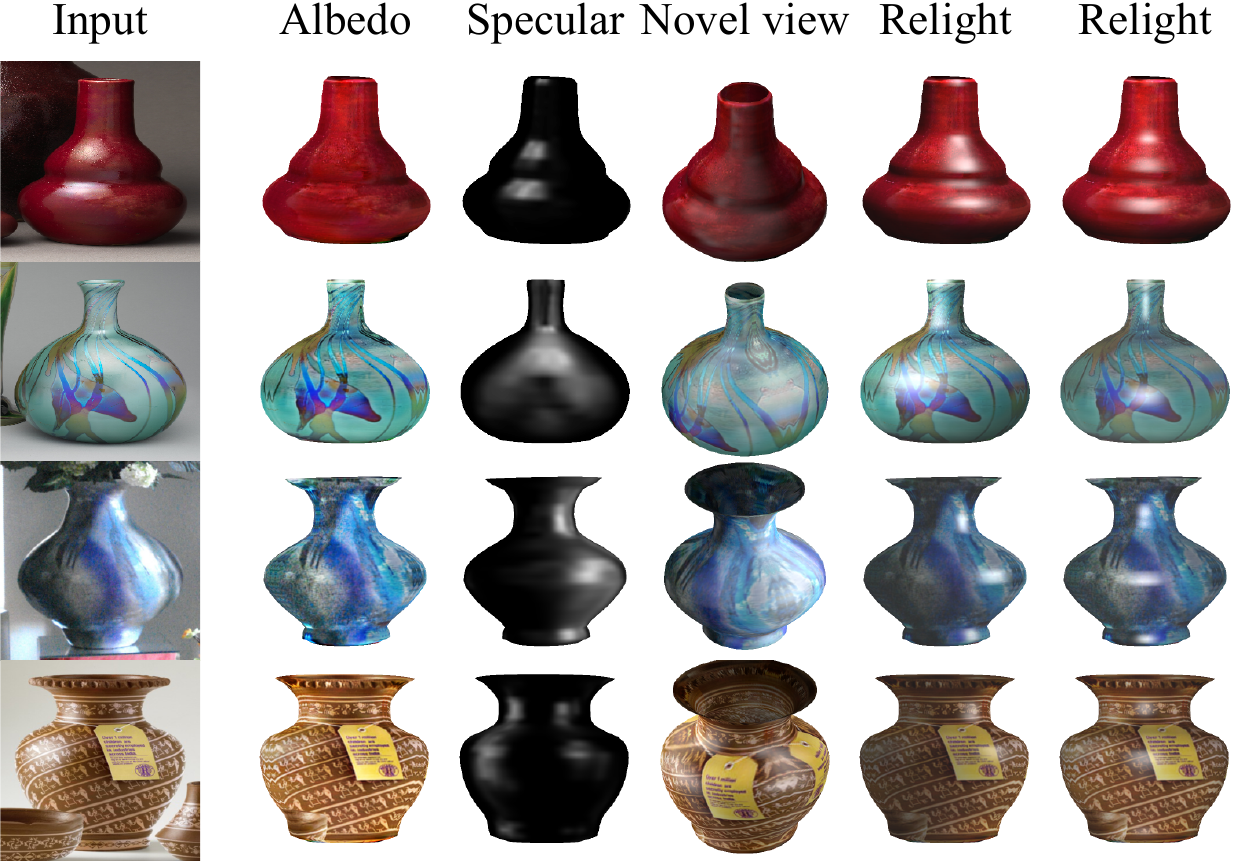}
\caption{\textbf{Novel view and relighting.}
Our method recovers accurate geometry and surface material, allowing us to render the vases from novel viewpoints and with new environment lighting.
Note that the last two examples are taken from Open Images~\cite{OpenImages}, which shows that
the model trained on museum images generalizes well to diverse input images beyond the training distribution.}
\label{fig:relight}
\vspace{-0.5em}
\end{figure}

\begin{table}[t]
\footnotesize\newcommand{\xpm}[1]{{\tiny$\pm#1$}}
\centering %
\begin{tabular}{lccc}
\toprule
  Method        & Albedo$^\mathsection$ {\tiny($\times 10^{-2}$)} $\downarrow$ & Normal$^\ddagger$ $\downarrow$ \\\midrule
  Ours          & $\mathbf{0.71}$ \xpm{0.92} & $\mathbf{5.81}$ \xpm{0.51} \\ \midrule
  ShapeNet-Intrinsics~\cite{ShapenetIntrinsics}      & $3.24$ \xpm{3.24} & - \\
  Single-Image SVBRDF~\cite{li2018learning}     & $3.34$ \xpm{2.48} & $36.39$ \xpm{6.92} \\
  SIRFS~\cite{BarronTPAMI2015}         & $2.74$ \xpm{3.28} & $35.85$ \xpm{11.15} \\
\bottomrule
\end{tabular}
\caption{\textbf{Quantitative comparison on synthetic vases.} We evaluate 
different methods quantitatively on our synthetic vase dataset.
Our method significantly outperforms other prior methods.
Error metrics: $^\mathsection$scale-invariant \textsc{mse}, following Grosse~\etal~\cite{grosse2009ground}, $^\ddagger$angular deviation in degrees.}
\label{tab:compare_syn}
\vspace{-0.5em}
\end{table}

\begin{table*}[t]
\footnotesize\newcommand{\xpm}[1]{{\tiny$\pm#1$}}
\centering %
\begin{tabular}{lccccccc}
\toprule
  Method                  & Pose$^\dagger$ $\downarrow$ & Normal$^\ddagger$ $\downarrow$ & Albedo$^\mathsection$ {\tiny($\times 10^{-2}$)} $\downarrow$ & Shininess$^\dagger$ $\downarrow$ & Spec. albedo$^\dagger$ $\downarrow$ & Env. map$^\mathsection$ $\downarrow$ \\\midrule
  Supervised              & $0.16$ \xpm{0.18} & $5.91$ \xpm{0.56} & $0.48$ \xpm{0.59} & $43.61$ \xpm{34.35} & $0.15$ \xpm{0.12} & $0.75$ \xpm{0.51} \\
  No decomposition        & - & - & $3.08$ \xpm{2.39} & - & - & - \\ \midrule
  Ours full               & $0.43$ \xpm{0.40} & $5.81$ \xpm{0.51} & $0.71$ \xpm{0.92} & $42.04$ \xpm{34.83} & $0.23$ \xpm{0.21} & $0.41$ \xpm{0.37} \\
  w/o $L_{\text{alb}}$    & $2.77$ \xpm{4.64} & $8.23$ \xpm{4.54} & $2.83$ \xpm{2.28} & $70.53$ \xpm{58.51} & $0.35$ \xpm{0.25} & $0.91$ \xpm{0.50} \\
  w/o $L_{\text{SAD}}$    & $0.48$ \xpm{0.49} & $5.83$ \xpm{0.48} & $0.75$ \xpm{1.09} & $43.57$ \xpm{33.59} & $0.21$ \xpm{0.18} & $0.41$ \xpm{0.39} \\
  w/o $L_{\text{diff}}$   & $0.45$ \xpm{0.42} & $5.78$ \xpm{0.49} & $0.82$ \xpm{0.98} & $43.27$ \xpm{34.00} & $0.33$ \xpm{0.21} & $0.46$ \xpm{0.35} \\
\bottomrule
\end{tabular}
\caption{\textbf{Baselines and ablations.}
We evaluate the predictions against the ground-truth on the synthetic vase dataset.
The performance of our model approaches the supervised baseline trained with full supervision, and the accuracy of the albedo prediction is clearly higher than lower-bound with no decomposition.
The ablation studies validate the effectiveness of each component.
Error metrics: $^\dagger$\textsc{rmse}, $^\ddagger$angular deviation in degrees, $^\mathsection$scale-invariant \textsc{mse}, following Grosse~\etal~\cite{grosse2009ground}.
}
\label{tab:ablation}
\vspace{-1em}
\end{table*}

\begin{figure}
\includegraphics[width=1\linewidth]{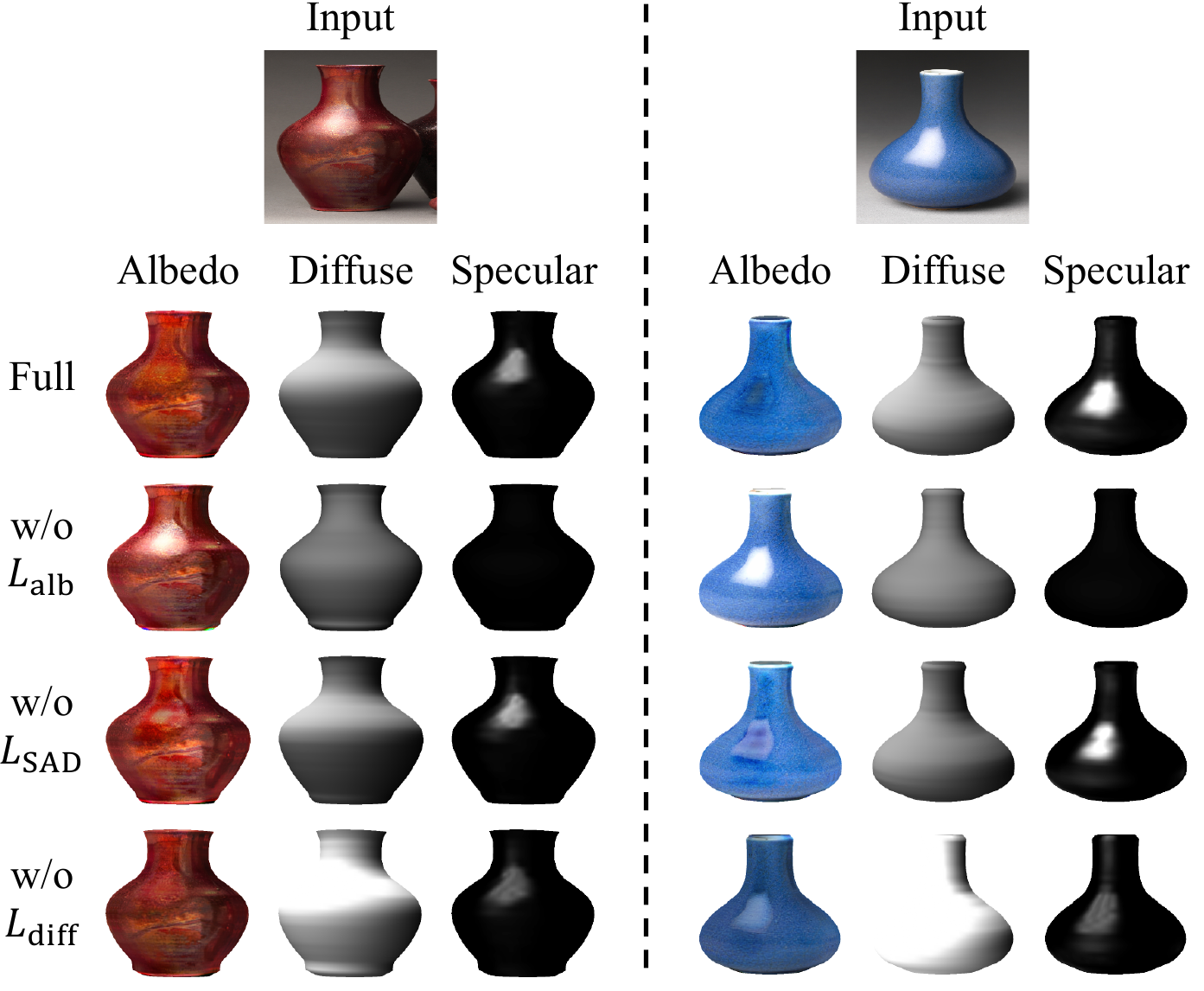}
\caption{\textbf{Qualitative ablation.}
We visualize the decomposition results of our full model and its variants.
Our full model recovers more realistic albedo and lighting compared to other variants.}
\label{fig:ablation}
\vspace{-1em}
\end{figure}

\subsection{Datasets and Implementation Details}
\label{sec:setup}
\noindent \textbf{Metropolitan Museum vases.}
We collected a dataset of real vase images from the Metropolitan Museum of Art Collection via the open-access API~\cite{metAPI}.
We first obtain 5,061 images with the query keyword ``vase'', and pass them through PointRend~\cite{kirillov2019pointrend} to generate bounding boxes and segmentation masks for each vase instance.
The images are then cropped and resized to $256 \times 256$, and GrabCut~\cite{rother2004grabcut} is applied to refine the masks.
We roughly filter out vases with non-SoR shapes manually, and split the remaining images into 1,888 training images and 526 testing images.

\myparagraph{Synthetic vases.}
Since we do not have ground-truth for the decomposed components on the real vases, in order to assess the performance of our model quantitatively, we prepare a dataset of synthetic vases.
We randomly generate vase-like SoR curves using combinations of sine curves, and take albedo maps from a public material dataset (CC0 Textures~\cite{cc0textures}) with various augmentations.
We then render the synthetic vases from random elevation angles in $(0,20^{\circ})$, assuming a Phong illumination model with random shininess values in $(1,200)$ and spherical Gaussian environment lighting~\cite{wang2009sglight,li2020inverse} with $3$ Gaussian lobes randomly sampled from the front upper hemisphere.
We generated 4,115 training images and 460 testing images.
See the supplementary material for some examples.

\myparagraph{Implementation details.}
The shape network $f_S$ consists of an encoder and two decoder branches.
The first decoder branch uses 1D upsampling convolutions to produce an $L \times 1$ radius column $\mathbf{\hat{r}}$, exploiting the structural prior of convolutions to obtain a smooth curve.
The second branch is simply $2$ FC layers that predict the height $\hat{h}$ and pose $\hat{v}$.
The albedo network $f_A$ is a U-Net~\cite{ronneberger2015u} with $6$ downsampling and $6$ upsampling layers.
The lighting network $f_L$ is similar to $f_S$, except that it predicts an environment map $\hat{E}$ with 2D upsampling convolutions, and a shininess scalar $\hat{\alpha}$ and a specular albedo scalar $\hat{\rho}$ with $2$ FC layers.
We use a Least Square GAN~\cite{mao2017squares} for \SAD, and the discriminator $D$ is a simple encoder network comprised of $5$ downsampling convolution layers.
All networks are trained with Adam with a learning rate of $0.0002$ and a batch size of $24$ for approximately $40$k iterations.

Both input images and unwrapped frontal texture maps are $256 \times 256$.
We use a projective camera with a narrow fixed field of view of $10^{\circ}$, since the images are cropped around the objects.
In practice, we only unwrap the frontal one third of the whole $360^{\circ}$ circular texture map to ignore the back of the vase and compensate for perspective projection.
The sizes of the vertex maps and the environment maps are $32 \times 96$ and $16 \times 48$ respectively.
For visualization, we replicate the texture maps three-fold, and use dimmed textures for the inside of the vase.
More details are included in the supplementary material.

\subsection{Qualitative Results on Real Vases}
Our method recovers geometry, specular material and lighting from a single image, and assumes no ground-truth labels except for object silhouettes during training.
To the best of our knowledge, no prior work tackles this problem under such a setting.
Nevertheless, we have identified several closest methods,
and show a comparison in \cref{fig:compare_real}.

SIRFS~\cite{BarronTPAMI2015} is an optimization-based method for decomposing albedo and diffuse shading from a single image, without considering specular materials.
ShapeNet-Intrinsics~\cite{ShapenetIntrinsics} predicts albedo, diffuse shading and specular shading from a single image without explicitly modeling lighting.
It is trained on synthetic ShapeNet objects with full supervision.
Single-Image SVBRDF~\cite{li2018learning} is another supervised method that predicts spatially-varying BRDF and environment lighting from a single input image, but assumes that images are captured under camera flash.
We also compare to Double-DIP~\cite{DoubleDIP}, an unsupervised method that decomposes a single image into multiple layers by exploiting the internal image statistics using ``Deep-Image-Prior'' networks~\cite{UlyanovVL17}, without requiring training data.
It achieves impressive decomposition results in several tasks, including reflection separation, motivating us to test it on the task of specularity separation.

Our method recovers accurate geometry and plausible disentanglement of material and lighting, whereas all other methods fail to decompose these components accurately.
Since ShapeNet-Intrinsics and Single-Image SVBRDF are trained with synthetic objects and real objects captured under camera flash respectively, they do not generalize well to these real vases under various lighting environments.
SIRFS results in poor decomposition in the presence of specularity, as it assumes only diffuse shading.
It is worth noting that Double-DIP in fact achieves plausible decomposition of albedo and specularity in scenarios where the surface textures are simple, although it fails when textures are more complicated. However, it does not consider 3D geometry and thus does not allow for realistic 3D editing.

\myparagraph{Novel views and relighting.}
Since our model recovers the 3D shape, surface material and lighting from a single image, we can easily render the object from arbitrary viewpoints under different lighting conditions, as shown in~\cref{fig:relight}.

\myparagraph{Generalization.} We further apply the trained model to diverse input images taken from the Open Images dataset~\cite{OpenImages}, shown in the last two row in~\cref{fig:relight}.
The model generalizes reasonably well to images beyond the training distribution of museum images, where the environment lighting may be more complicated or the vase may be partially occluded.

\subsection{Quantitative Comparisons on Synthetic Data}
To quantify the prediction accuracy, we evaluate it on our synthetic vase dataset, and report a numerical comparison of different methods in~\cref{tab:compare_syn}.
We measure the accuracy of the albedo in the predicted region using a scale-invariant mean square error metric~\cite{grosse2009ground}, since the scales of the albedo intensity and the lighting intensity are ambiguous (one can trade off one for the other),
and measure the accuracy of normal maps in degrees of angular deviation.
It is evident in~\cref{tab:compare_syn} that our method outperforms other methods for both albedo and shape predictions.

\subsection{Baselines and Ablations} \label{sec:ablation}
We conduct a thorough evaluation of all the predictions of our model on the synthetic dataset and compare the results with two baselines and various ablated models in~\cref{tab:ablation,fig:ablation}.
The first baseline is a supervised model trained with ground-truth labels on all predictions, which gives an performance upper-bound.
We also report a performance lower-bound on the albedo decomposition, obtained by simply evaluating the albedo error metric on the input image without any decomposition.
The error of our predicted albedo is clearly much lower than the original undecomposed image, and our model overall achieves high performance close to the supervised baseline on all metrics.

Comparing our full model to the ablated models, it is evident that without the single-color albedo rendering loss $L_{\text{alb}}$, the model fails to learn various components.
The diffuse regularizer $L_{\text{diff}}$ largely improves the lighting prediction and consequently other predicted components as well.
The albedo discriminator loss $L_{\text{SAD}}$ also improves accuracy of albedo prediction,
and more importantly, it helps inpaint the albedo in the specular regions as visualized in~\cref{fig:ablation}.

\section{Conclusion}

\begin{figure}
\includegraphics[width=1\linewidth]{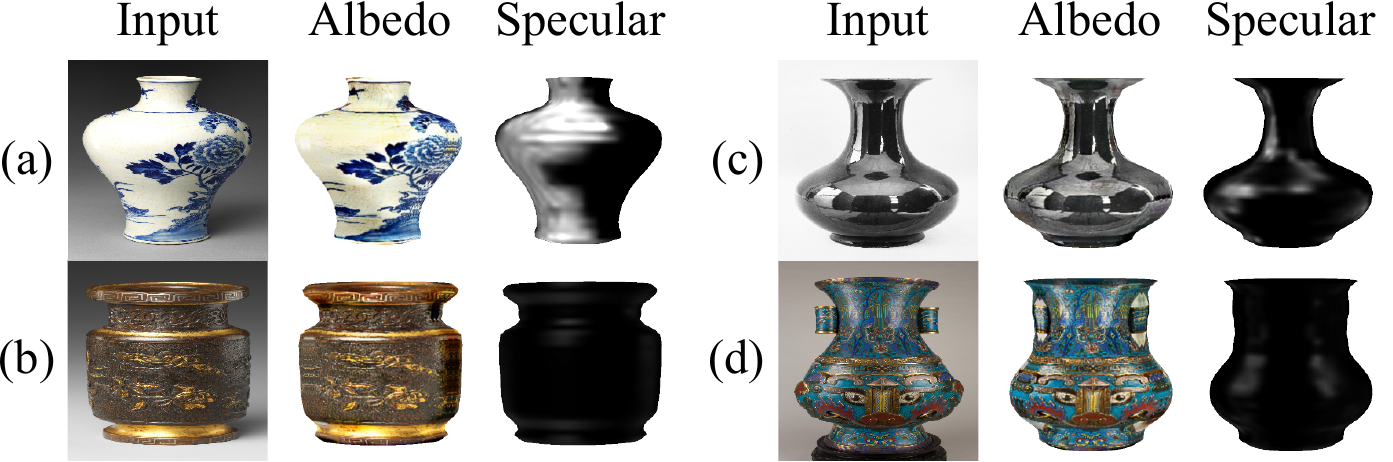}
\caption{\textbf{Limitations.}
(a) Incorrect environment lighting and specularity prediction in the presence of high contrast textures.
(b) Spatially-varying material properties.
(c) Complicated environment lighting.
(d) Non-revolutionary shapes.}
\label{fig:failure}
\vspace{-0.5em}
\end{figure}

We introduce an end-to-end framework for de-rendering a single image into shape, lighting, and surface material components, learning only from single-image collections with 2D silhouettes. Our method works well on both synthetic and real images of revolutionary artefacts and enables applications such as free-view rendering and relighting.

\myparagraph{Limitations and future work.} 
\cref{fig:failure} illustrates limitations of our method. 
First, it tends to predict specularity in bright texture regions, which could lead to unrealistic environment lighting in the presence of high-contrast textures.
This could be improved by adding constraints on the lighting model. 
Second, since we use a Phong 
model with a single shininess constant for each vase and a low-resolution environment illumination map, our model cannot handle objects with spatially-varying material properties or complex lighting. 
We intend to incorporate more sophisticated graphics models in future work.
Last, as a first step to tackle this extremely challenging problem, we assume revolutionary objects, and hence our model does not work well on objects whose shapes are not revolutionary. 
However, the proposed components for disentangling lighting and albedo, including the self-supervised discriminator, are not specific to revolutionary objects and it would be interesting to extend these ideas to general real-world objects.

\paragraph{Acknowledgements}
We would like to thank Christian Rupprecht, Soumyadip Sengupta, Manmohan Chandraker and Andrea Vedaldi for insightful discussions.

{\small
\bibliographystyle{ieee_fullname}
\bibliography{ref}
}

\newpage

\section{Supplementary Material}

\subsection{Training Details}
All hyper-parameter settings are specified in~\cref{tab:params} and the network architectures in~\cref{tab:arch_shape,tab:arch_light,tab:arch_albedo,tab:arch_disc}.
Abbreviations of the components are defined as follows:

\begin{itemize}
	\item $\text{Conv}(c_{in}, c_{out}, k, s, p)$: 2D convolution with $c_{in}$ input channels, $c_{out}$ output channels, kernel size $k$, stride $s$ and padding $p$.
	\item $\text{Deconv}(c_{in}, c_{out}, k, s, p)$: 2D deconvolution with $c_{in}$ input channels, $c_{out}$ output channels, kernel size $k$, stride $s$ and padding $p$.
	\item $\text{Upsample}(s)$: 2D nearest-neighbor upsampling with a scale factor $s$.
	\item $\text{Linear}(c_{in}, c_{out})$: linear layer with $c_{in}$ input channels and $c_{out}$ output channels.
	\item $\text{GN}(n)$: group normalization~\cite{GroupNorm2018}.
	\item $\text{IN}$: instance normalization~\cite{ulyanov2016instance} with $n$ groups.
	\item $\text{LReLU}(p)$: leaky ReLU~\cite{lrelu2013} with a slope $p$.
	\item Conv1D and Upsample1D are similarly defined.
\end{itemize}

\subsection{Synthetic Vases}
\begin{figure}
\includegraphics[width=1\linewidth]{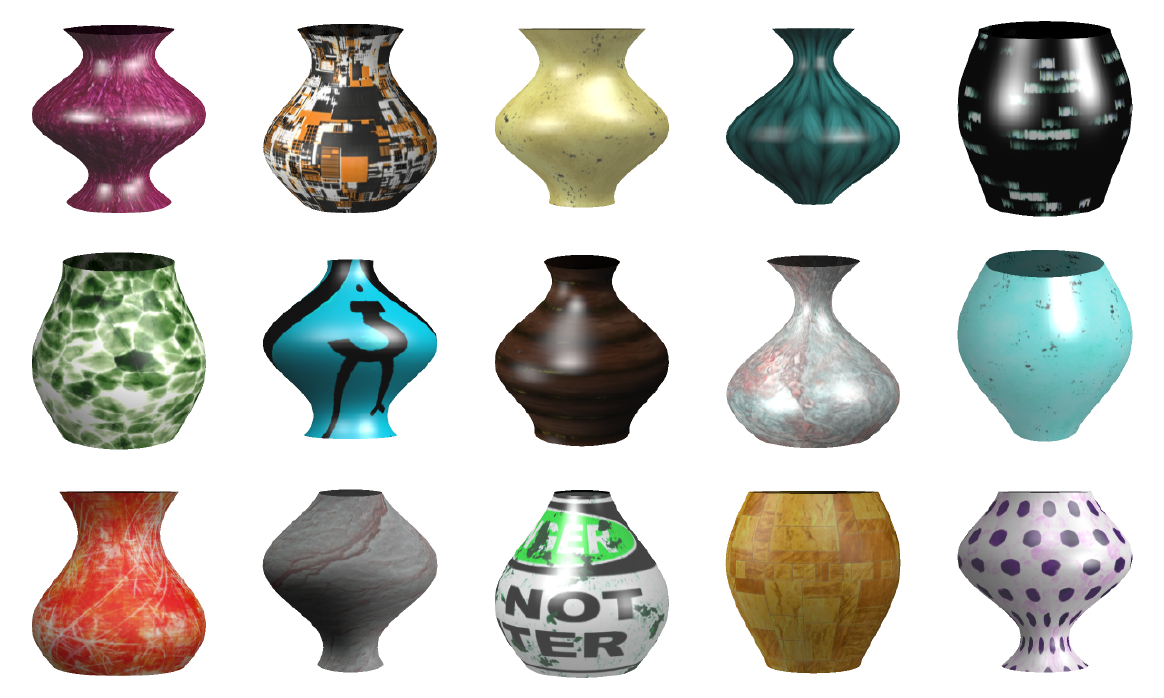}
\caption{
Examples of the synthetic vases.}
\label{fig:syn_vase}
\end{figure}

We generate a synthetic vase dataset in order to conduct quantitative assessment of our de-rendering results.
Examples of the synthetic vases are shown in~\cref{fig:syn_vase}.
The detailed procedure to generate this dataset is described in the following.

\paragraph{SoR shapes.}
We simulate vase-like SoR curves $\mathbf{r} \in \mathbb{R}^L$ using a combination of two sine curves, where $L$ is set to be $32$, and each entry $r_i$ is given by:
\begin{equation}
\begin{aligned}
    r_i &= t + f_1(i) + f_2(i) \\
    f_1(i) &= a_1 \cdot (1 + \sin(\frac{L-i}{L} \cdot p_1 + \frac{i}{L} \cdot q_1)) \\
    f_2(i) &= a_2 \cdot (1 + \sin(p_2 + \frac{i}{L} \cdot q_2)),
\end{aligned}
\end{equation}
where the random variables are $t \sim \mathcal{U}(0.1, 0.3)$, $a_1 \sim \mathcal{U}(0, 0.3)$, $p_1 \sim \mathcal{U}(-\pi, 0)$, $q_1 \sim \mathcal{U}(\frac{\pi}{2}, 2\pi)$, $a_2 \sim \mathcal{U}(0, 0.1)$, $p_2 \sim \mathcal{U}(0, 2\pi)$ and $q_2 \sim \mathcal{U}(\frac{\pi}{2}, 2\pi)$.

We then render the vases with random elevation angles between $0^\circ$ and $20^\circ$, using a projective camera with a field of view of $10^\circ$.

\paragraph{Material.}
We generate random diffuse albedo maps using texture images from a public material dataset (CC0 Textures~\cite{cc0textures}), with random augmentations in brightness, contrast and hue.
Shininess constant $\alpha$ is randomly sampled between $1$ and $196$ and specular albedo constant $\rho$ is sampled between $0.1$ and $1$.

\begin{table}[t]
\footnotesize
\begin{center}
\begin{tabular}{lc}
\toprule
 Parameter & Value/Range \\ \midrule
 Optimizer & Adam \\
 Learning rate & $2\times 10^{-4}$ \\
 Number of iterations & $40$k \\
 Batch size & $24$ \\
 Loss weight $\lambda_{\text{s}}$ & $10$ \\
 Loss weight $\lambda_{\text{dt}}$ & $100$ \\
 Loss weight $\lambda_{\text{im}}$ & $1$ \\
 Loss weight $\lambda_{\text{alb}}$ & $1$ \\
 Loss weight $\lambda_{\text{SAD}}$ & $0.01$ \\
 Loss weight $\lambda_{\text{diff}}$ & $1$ \\ \midrule
 Input image size & $256 \times 256$ \\
 Whole unwrapped image size & $256 \times 768$ \\
 Frontal unwrapped image size & $256 \times 256$ \\
 Vertex grid size & $32 \times 96$ \\
 Environment map size & $16 \times 48$ \\
 Field of view (FOV) & $10^\circ$ \\
 Radius $\hat{\mathbf{r}}$ & $(0.05, 0.9)$ \\
 Radius column height $\hat{h}$ & $(0.5, 0.95)$ \\
 Pitch angles & $(0^\circ, 20^\circ)$ \\
 Roll angles & $(-10^\circ, 10^\circ)$ \\
 Translation in $X$, $Y$ axes & $(-0.2, 0.2)$ \\
 Albedo $\hat{A}$ & $(0, 1)$ \\
 Shininess $\hat{\alpha}$ & $(1, 196)$ \\
 Specular albedo $\hat{\rho}$ & $(0, 2)$ \\
 Environment map $\hat{E}$ & $(0, 1)$ \\
\bottomrule
\end{tabular}
\end{center}
\caption{Training details and hyper-parameter settings.}\label{tab:params}
\end{table}
\begin{table}[t]
\footnotesize
\begin{center}
\begin{tabular}{lc}
\toprule
 Encoder & Output size \\ \midrule
 Conv(3, 64, 4, 2, 1) + ReLU & 128 $\times$ 128\\
 Conv(64, 128, 4, 2, 1) + ReLU & 64 $\times$ 64\\
 Conv(128, 256, 4, 2, 1) + ReLU & 32 $\times$ 32\\
 Conv(256, 512, 4, 2, 1) + ReLU & 16 $\times$ 16\\
 Conv(512, 512, 4, 2, 1) + ReLU & 8 $\times$ 8\\
 Conv(512, 512, 4, 2, 1) + ReLU & 4 $\times$ 4\\
 Conv(512, 128, 4, 1, 0) + ReLU & 1 $\times$ 1\\ \midrule \midrule
 Decoder & Output size \\ \midrule
 Upsample1D(2) + Conv1D(128, 128, 3, 1, 1) + ReLU & 2\\
 Upsample1D(2) + Conv1D(128, 128, 3, 1, 1) + ReLU & 4\\
 Upsample1D(2) + Conv1D(128, 128, 3, 1, 1) + ReLU & 8\\
 Upsample1D(2) + Conv1D(128, 128, 3, 1, 1) + ReLU & 16\\
 Upsample1D(2) + Conv1D(128, 128, 3, 1, 1) & 32\\
 \enskip \rotatebox[origin=c]{180}{$\Lsh$} Sigmoid $\rightarrow$ output $\hat{\mathbf{r}}$ & 32\\
 Linear(128, 128) + ReLU & 1\\
 Linear(128, 5) & 1\\
 Sigmoid $\rightarrow$ output $\hat{h}, \hat{v}$ & 1\\
\bottomrule
\end{tabular}
\end{center}
\caption{Architecture of the shape network $f_S$. The network outputs radius column $\hat{\mathbf{r}}$, height $\hat{h}$ and camera pose $\hat{v}$ from two branches.}

\label{tab:arch_shape}
\end{table}
\begin{table}[t]
\footnotesize
\begin{center}
\resizebox{.47\textwidth}{!}{
\begin{tabular}{lc}
\toprule
 Encoder & Output size \\ \midrule
 Conv(3, 64, 4, 2, 1) + GN(16) + LReLU(0.2) & 128 $\times$ 128\\
 Conv(64, 128, 4, 2, 1) + GN(32) + LReLU(0.2) & 64 $\times$ 64\\
 Conv(128, 256, 4, 2, 1) + GN(64) + LReLU(0.2) & 32 $\times$ 32\\
 Conv(256, 512, 4, 2, 1) + GN(128) + LReLU(0.2) & 16 $\times$ 16\\
 Conv(512, 512, 4, 2, 1) + GN(128) + LReLU(0.2) & 8 $\times$ 8\\
 Conv(512, 512, 4, 2, 1) + LReLU(0.2) & 4 $\times$ 4\\
 Conv(512, 128, 4, 1, 0) + ReLU & 1 $\times$ 1\\ \midrule \midrule
 Decoder & Output size \\ \midrule
 Deconv(128, 512, (2,6), 1, 0) + ReLU & 2 $\times$ 6\\
 Upsample(2) + Conv(512, 256, 3, 1, 1) + GN(64) + ReLU & 4 $\times$ 12\\
 Upsample(2) + Conv(256, 128, 3, 1, 1) + GN(32) + ReLU & 8 $\times$ 24\\
 Upsample(2) + Conv(128, 64, 3, 1, 1) + GN(16) & 16 $\times$ 48\\
 \enskip \rotatebox[origin=c]{180}{$\Lsh$} Sigmoid $\rightarrow$ output $\hat{E}$ & 16 $\times$ 48\\
 Linear(128, 128) + ReLU & 1\\
 Linear(128, 2) & 1\\
 Sigmoid $\rightarrow$ output $\hat{\alpha}, \hat{\rho}$ & 1\\
\bottomrule
\end{tabular}
}
\end{center}
\caption{Architecture of the light network $f_L$. The network outputs environment map $\hat{E}$ and specular albedo $\hat{\rho}$ from two branches.}

\label{tab:arch_light}
\end{table}
\begin{table}[t]
\footnotesize
\begin{center}
\resizebox{.47\textwidth}{!}{
\begin{tabular}{lc}
\toprule
 Encoder & Output size \\ \midrule
 Conv(3, 64, 4, 2, 1) + IN + LReLU(0.2) & 128 $\times$ 128\\
 Conv(64, 128, 4, 2, 1) + IN + LReLU(0.2) & 64 $\times$ 64\\
 Conv(128, 256, 4, 2, 1) + IN + LReLU(0.2) & 32 $\times$ 32\\
 Conv(256, 512, 4, 2, 1) + IN + LReLU(0.2) & 16 $\times$ 16\\
 Conv(512, 512, 4, 2, 1) + IN + LReLU(0.2) & 8 $\times$ 8\\
 Conv(512, 512, 4, 2, 1) + IN + LReLU(0.2) & 4 $\times$ 4\\ \midrule \midrule
 Decoder & Output size \\ \midrule
 Upsample(2) + Conv(512, 512, 3, 1, 1) + IN + SC + ReLU & 8 $\times$ 8\\
 Upsample(2) + Conv(512, 256, 3, 1, 1) + IN + SC + ReLU & 16 $\times$ 16\\
 Upsample(2) + Conv(512, 256, 3, 1, 1) + IN + SC + ReLU & 32 $\times$ 32\\
 Upsample(2) + Conv(256, 128, 3, 1, 1) + IN + SC + ReLU & 64 $\times$ 64\\
 Upsample(2) + Conv(128, 64, 3, 1, 1) + IN + SC + ReLU & 128 $\times$ 128\\
 Upsample(2) + Conv(64, 3, 3, 1, 1) & 256 $\times$ 256\\
 Tanh $\rightarrow$ output $\hat{A}$ & 256 $\times$ 256\\
\bottomrule
\end{tabular}
}
\end{center}
\caption{Architecture of the albedo network $f_A$. The network follows a U-Net structure with skip-connections and replaces deconvolution with nearest neighbor upsampling followed by convolution.}

\label{tab:arch_albedo}
\end{table}
\begin{table}[t]
\footnotesize
\begin{center}
\begin{tabular}{lc}
\toprule
 Encoder & Output size \\ \midrule
 Conv(3, 64, 4, 2, 1) + IN + LReLU(0.2) & 32 $\times$ 32\\
 Conv(64, 128, 4, 2, 1) + IN + LReLU(0.2) & 16 $\times$ 16\\
 Conv(128, 256, 4, 2, 1) + IN + LReLU(0.2) & 8 $\times$ 8\\
 Conv(256, 512, 4, 2, 1) + LReLU(0.2) & 4 $\times$ 4\\
 Conv(512, 1, 4, 1, 0) $\rightarrow$ output scalar & 1 $\times$ 1\\
\bottomrule
\end{tabular}
\end{center}
\caption{Architecture of the discriminator network $D$. The network outputs a single scalar for each input patch.}

\label{tab:arch_disc}
\end{table}

\begin{figure*}
\begin{center}
\includegraphics[width=0.9\linewidth]{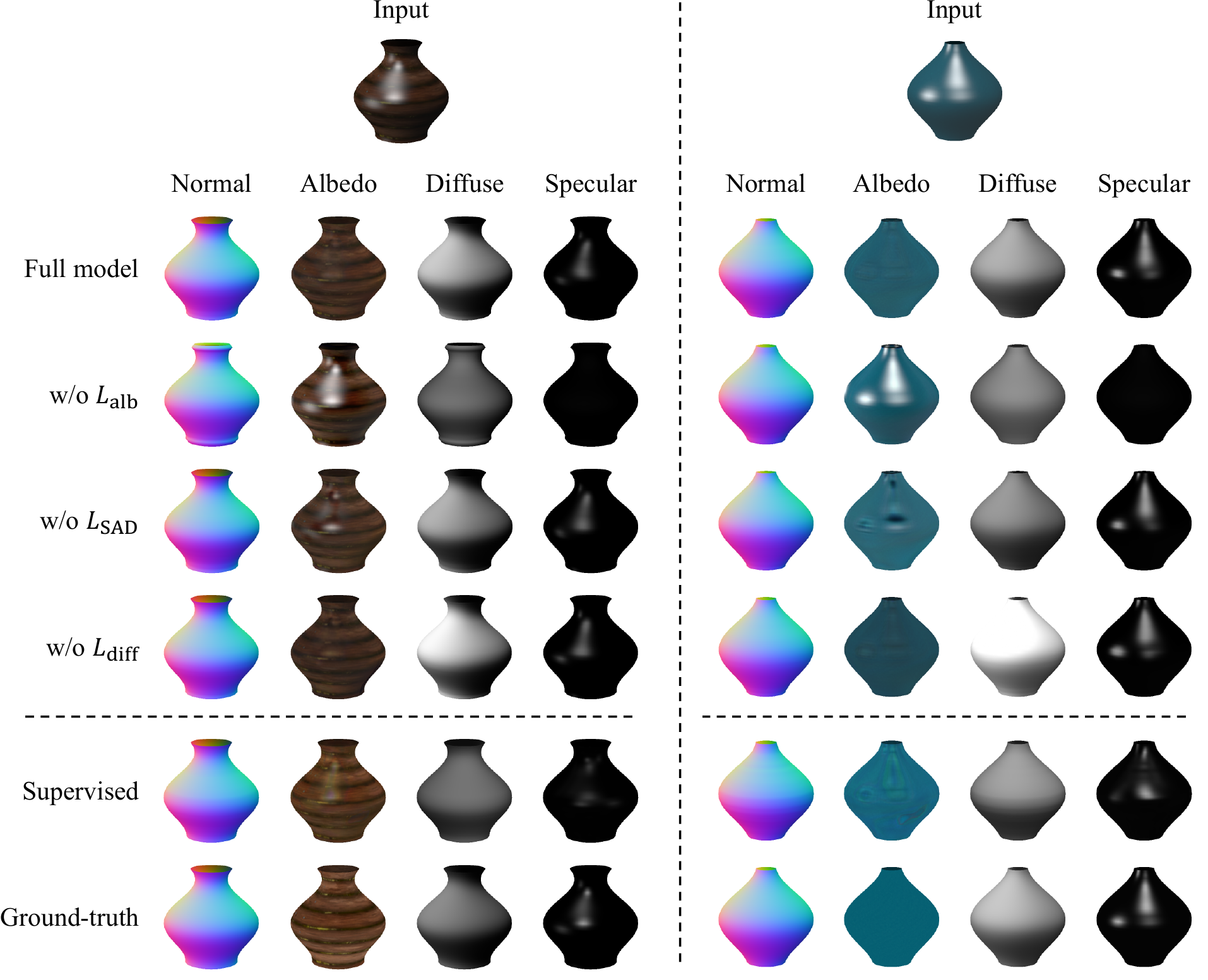}
\caption{
Qualitative comparison of the ablation experiments and the supervised baseline.}
\label{fig:ablation_syn}
\end{center}
\end{figure*}

\paragraph{Lighting.}
We synthesize environment illumination using $3$ random spherical Gaussian lobes~\cite{wang2009sglight,li2020inverse}:
\begin{equation}
    L(\eta) = \sum_{k=1}^3 \sqrt{\lambda_k} F_k G(\eta; \xi_k, \lambda_k), \,\,
    G(\eta; \xi, \lambda) = e^{-\lambda(1-\eta \cdot \xi)},
\end{equation}
where $\xi_k$ controls the direction of each lobe and is a unit vector randomly sampled from the upper-front quarter of the sphere, $\lambda_k \sim \mathcal{U}(10, 30)$ controls the bandwidth, and $F_k \sim \mathcal{U}(0.1, 0.3)$ controls the intensity.

\subsection{Additional Results}

\cref{fig:ablation_syn} shows a visual comparison of the results obtained from the ablation experiments as well as a supervised baseline, corresponding to the numerical results reported in~\cref{tab:ablation}.

\begin{figure*}
\begin{center}
\includegraphics[width=0.74\linewidth]{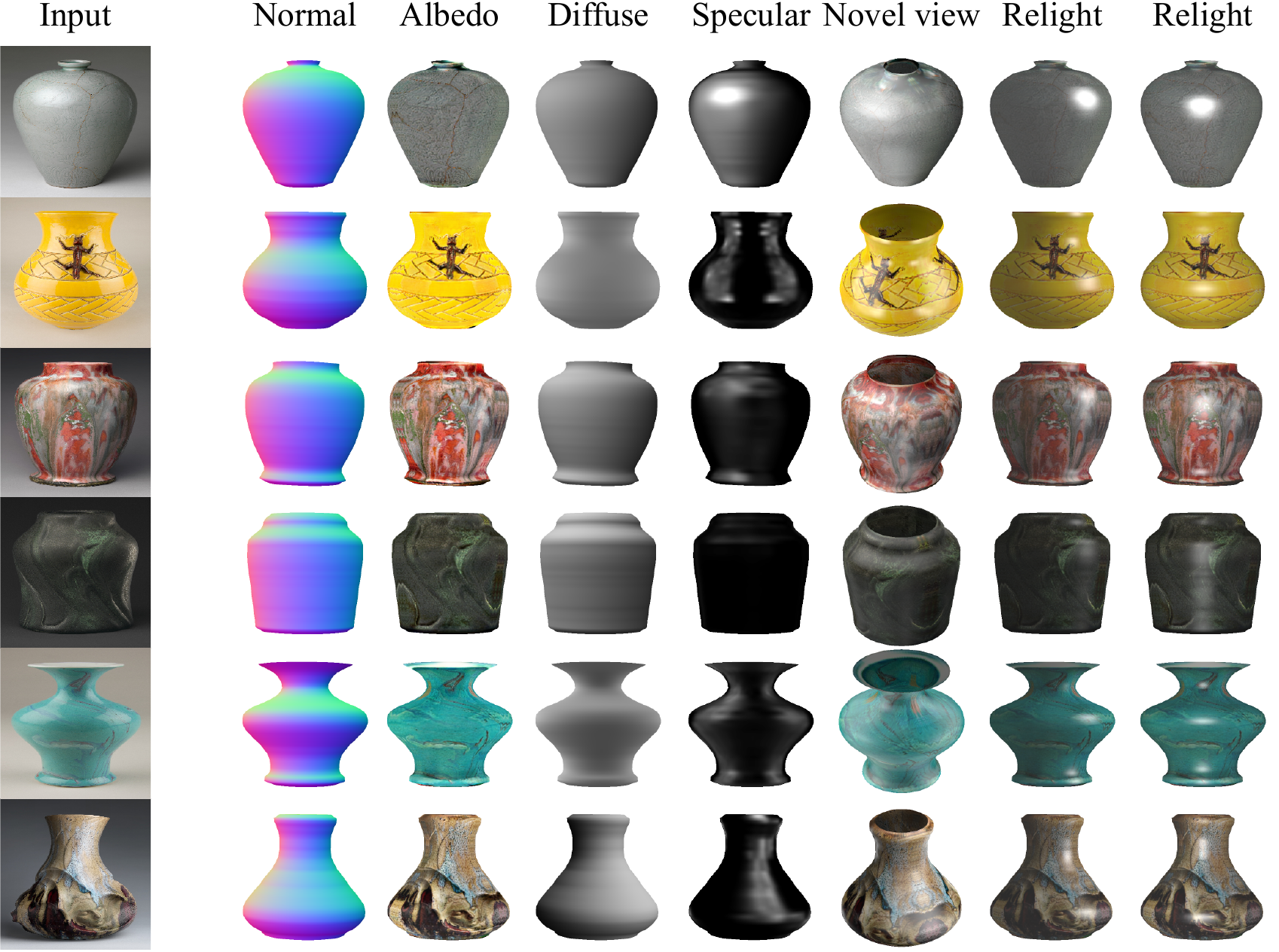}
\caption{
Additional results on Metropolitan Museum collection~\cite{metAPI}.}
\label{fig:supmat_met_vase}
\end{center}
\end{figure*}

\begin{figure*}
\begin{center}
\includegraphics[width=0.74\linewidth]{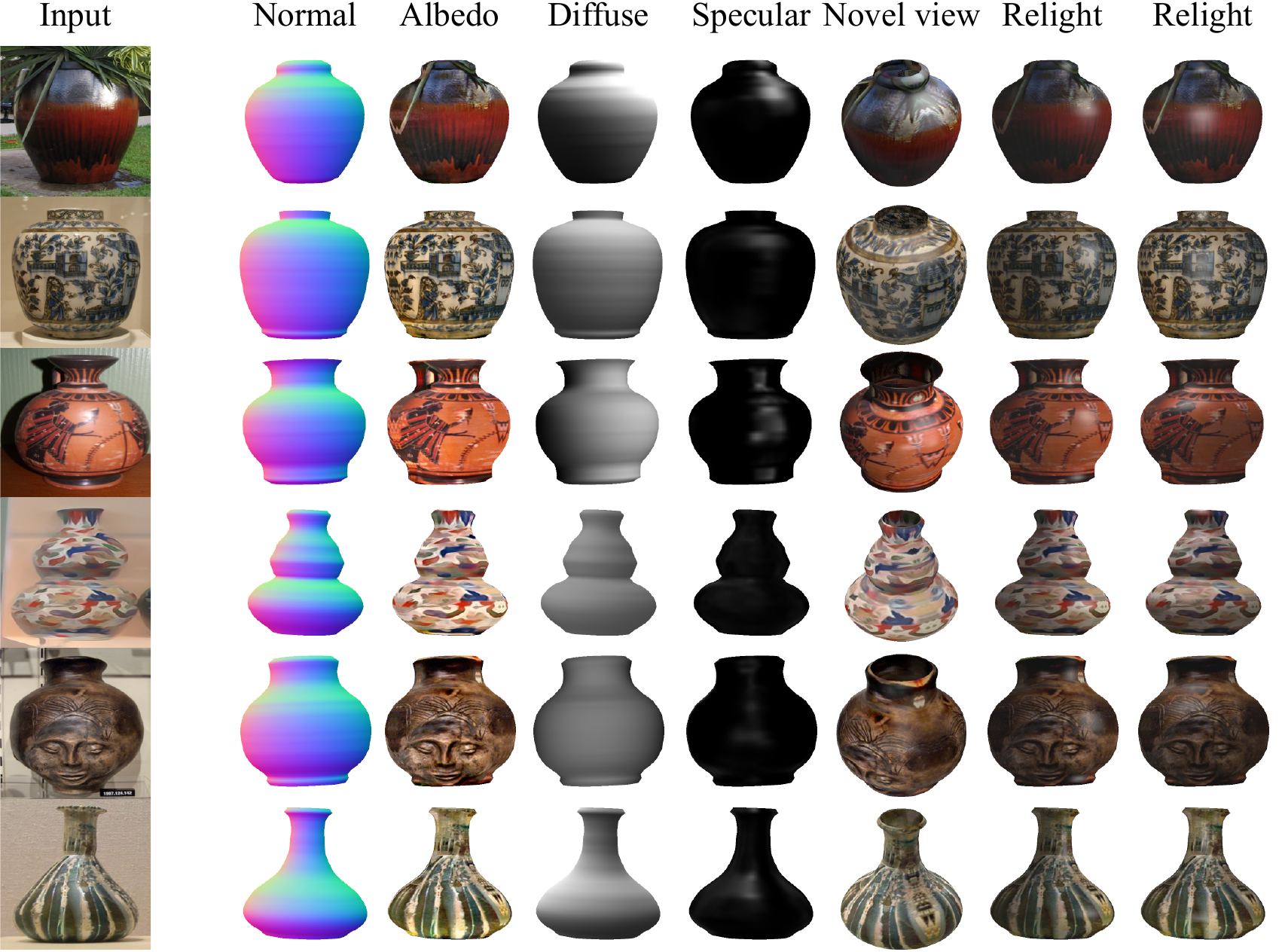}
\caption{
Additional results on Open Images vases~\cite{OpenImages}.}
\label{fig:supmat_other_vase}
\end{center}
\end{figure*}

Additional decomposition and relighting results of real vases are shown in~\cref{fig:supmat_met_vase} (from Metropolitan Museum collection~\cite{metAPI}) and in~\cref{fig:supmat_other_vase} (from Open Images~\cite{OpenImages}).
See the video for more visual results, including animations of rotating vases as well as relighting effects.

\end{document}